\documentclass{IEEEtran}
\usepackage{edgeStd}

\usepackage{amsthm}
\crefformat{equation}{(#2#1#3)}
\Crefformat{equation}{(#2#1#3)}

\allowdisplaybreaks
\newcommand{\unif}{\text{Unif}}
\newcommand{\is}{\text{IS}}

\begin{document}
\title{Max-Entropy Reinforcement Learning with Flow Matching and A Case Study on LQR}
\author{Yuyang Zhang$^{1,2}$, Yang Hu$^1$, Bo Dai$^3$, and Na Li$^{1}$
\thanks{This work is supported by NSF AI institute 2112085, NSF
ECCS 2328241, NIH R01LM014465. Yuyang Zhang is supported by the Kempner Graduate Fellowship.}
\thanks{$^{1}$Yuyang Zhang, Yang Hu, and Na Li are with the School of Engineering and Applied Sciences, Harvard University, USA.
{\tt\small \{yuyangzhang@g, yanghu@g, nali@seas\}.harvard.edu.}
}%
\thanks{$^{2}$Yuyang Zhang is also with the Kempner Institute, Harvard University.}
\thanks{$^{3}$Bo Dai is with the College of Computing, Georgia Institute of Technology, USA.
{\tt\small bodai@cc.gatech.edu}
}%
}

\maketitle

\begin{abstract}
    Soft actor-critic (SAC) is a popular algorithm for max-entropy reinforcement learning. In practice, the energy-based policies in SAC are often approximated using simple policy classes for efficiency, sacrificing the expressiveness and robustness. 
    In this paper, we propose a variant of the SAC algorithm that parameterizes the policy with flow-based models, leveraging their rich expressiveness. In the algorithm, we evaluate the flow-based policy utilizing the instantaneous change-of-variable technique and update the policy with an online variant of flow matching developed in this paper. This online variant, termed importance sampling flow matching (ISFM), enables policy update with only samples from a user-specified sampling distribution rather than the unknown target distribution. We develop a theoretical analysis of ISFM, characterizing how different choices of sampling distributions affect the learning efficiency. Finally, we conduct a case study of our algorithm on the max-entropy linear quadratic regulator problems, demonstrating that the proposed algorithm learns the optimal action distribution. 
\end{abstract}


\section{Introduction}
Maximum entropy reinforcement learning (max-entropy RL) has emerged as a popular reinforcement learning (RL) framework for a wide spectrum of tasks \cite{maxent_rl1,SAC,maxent_rl2}. Based on the max entropy principle, it optimizes a return function augmented by entropy regularization of the current policy. Empirical studies have demonstrated that max-entropy RL learns policies that not only maximize rewards, but are also robust to the environmental disturbances \cite{maxent_rl3}. 

To solve max-entropy RL problems, one of the most prominent algorithms is Soft Actor-Critic (SAC) \cite{SAC}. It maintains an energy-based model as its policy, and alternates between policy improvement and policy evaluation to improve the entropy-regularized return. Theoretical analysis has shown that SAC indeed converges to the optimal solution of max-entropy RL given that the policy evaluation and improvement steps are performed accurately \cite{SAC, SAC_theory}. 
However, the improvement and evaluation of energy-based policies remain notoriously challenging. Early MCMC-based algorithms improve the energy-based policy by learning the gradient of an energy function \cite{ebm_1}. Their policy improvement has limited accuracy in RL, because the required gradient information is difficult to approximate. Recent work uses diffusion models to parameterize the energy-based policies \cite{diffusion_1,diffusion_2,diffusion_3}. Their evaluation is intractable due to the lack of efficient methods to approximate policy entropy.
In practice, many existing algorithms restrict the policy class to simple families, such as Gaussian policies \cite{SAC,gaussian_1}. This restriction significantly limits the expressiveness and robustness of the learned policies. 

Fortunately, flow-based models have emerged as powerful universal distribution approximators \cite{flow_approximation_1,flow_approximation_2,flow_approximation_3}. To train flow-based models, a popular and efficient algorithm is flow matching, which matches the velocity field of the flow using samples from the target distribution.
Unlike traditional likelihood-based training methods \cite{flow_approximation_2,nf_1}, flow matching neither involves Jacobian gradient calculation, nor constrains the neural network structure to be invertible, which makes it more computationally efficient and flexible \cite{prelim_flow}.
These make flow-based models a promising choice to parameterize energy-based policies in max-entropy RL. 

In this paper, we propose to parameterize policies in the SAC algorithm with flow-based models trained via flow matching. 
\textit{For policy improvement}, we propose an online variant of flow matching, termed importance sampling flow matching (ISFM). In contrast to standard flow matching that requires samples from the unknown target distribution, ISFM requires only samples from a user-specified distribution, making it appropriate for online RL settings. We develop a theoretical analysis of ISFM, characterizing how different choices of sampling distributions affect the learning accuracy. We then use the result to guide the choice of sampling distributions. \textit{For policy evaluation,} we leverage the instantaneous change-of-variable formula \cite{prelim_ode_intro}, a well-established result in flow-based model literature, to evaluate the entropy-regularized return of the flow-based policy. 
\textit{Finally, as a case study for the proposed algorithm,} we investigate the max-entropy Linear Quadratic Regulator (LQR) problems \cite{maxent_lqr}. We show that the SAC algorithm, integrated with flow-based policies, learns the optimal solution. 

The rest of the paper is organized as follows. In \Cref{sec:prelim}, we introduce necessary background on max-entropy RL and flow matching. In \Cref{sec:alg}, we propose the adapted SAC algorithm framework, providing details on policy evaluation and improvement for flow-based policies. Finally, in \Cref{sec:lqr}, we finish the paper with a case study on max-entropy LQR problems, including both theoretical and simulation justifications of the proposed algorithm. For clarity of the paper, proofs of the theoretical results are deferred to Appendix \ref{sec:proof}.

\section{Preliminaries}\label{sec:prelim}
\subsection{Max-Entropy Reinforcement Learning}
An MDP can be defined by the tuple $(\calX, \calU, \rho^0, P, r, \gamma)$, where $\calX$ is the state space, $\calU$ is the action space, $\rho^0$ is the initial state distribution, $P: \calX\times \calU\mapsto \Delta(\calU)$ is the transition probability, $r: \calX\times\calU\mapsto \bbR$ is the reward function, and $\gamma$ is the discount factor. 

Consider any policy $\pi: \calX\mapsto\Delta(\calU)$ that maps a state to an action distribution. We define the corresponding entropy-regularized return function: 
\begin{equation*}\begin{split}
    J^{\pi}(x) = \bbE\bs{\sum_{t=0}^{\infty} \gamma^{t}\bs{r(x^t,u^t) + \alpha\calH(\pi(\cdot|x^t))}\big| x^0=x},
\end{split}\end{equation*}
where superscript $t$ denotes the time step in MDP.
In max-entropy RL, our objective is to find the optimal policy $\pi^\star$ that maximizes $\bbE_{x\sim\rho^0}J^{\pi}(x)$.

To facilitate later discussions, we also define the soft Q function of policy $\pi$:
\begin{equation*}\begin{split}
    {}& Q^{\pi}(x,u) \coloneqq r(x,u) + \\
    {}& \bbE\left[\sum_{t=1}^\infty \gamma^t \bs{r(x^t,u^t) + \alpha\calH\b{\pi(\cdot|x^t)}}\Big|(x^0,u^0)=(x,u)\right].
\end{split}\end{equation*}
It is the only fixed point of the following soft Bellman operator $\calT^{\pi}$ \cite{SAC}:
\begin{equation}\begin{split}\label{eq:softQ}
    {}& \calT^{\pi}Q(x,u) \coloneqq r(x,u)\\
    + {}& \gamma\bbE_{x'\sim\bbP(x,u),u'\sim\pi(\cdot|x')}\bs{Q(x',u') + \alpha\calH(\pi(\cdot|x'))}.
\end{split}\end{equation}

\subsection{Flow Matching}
Flow-based models are expressive models for learning distributions. They define a time-dependent vector field $v: \bbR^{d}\times [0,1]\mapsto \bbR^d$ that governs the evolution of random variable $u_{\tau}$ through the oridinary differential equation (ODE) $\d u_\tau/\d \tau=v_\tau(u_\tau)$. Here subscript $\tau\in[0,1]$ denotes the time in the ODE. Under the ODE, samples $u_0\sim p_0$ are transformed to samples $u_\tau\sim p_\tau, \forall \tau\in[0,1]$. We say that $p_{\tau}$ is generated by vector field $v_{\tau}$. 

The vector field transforming $p_0$ to $p_1$ is generally not unique. To specify a vector field that facilitates learning, we first choose a conditional probability path $p_\tau(u_\tau|u_1)$ such that it is easy to derive a conditional vector field $v_{\tau}(u_{\tau}|u_1)$ generating $p_{\tau}(u_{\tau}|u_1)$. Moreover, $p_{\tau}(u_{\tau}|u_1)$ should be consistent with $p_0(u_0)$ and $p_1(u_1)$, in the sense that $\int p_0(u|u_1) p_1(u_1)\d u_1 = p_0(u)$ and $\int p_1(u|u_1) p_1(u_1)\d u_1 = p_1(u)$. 
The corresponding marginal probability path is then given by $p_{\tau}(u_{\tau}) = \int p_{\tau}(u_\tau|u_1)p_1(u_1)\d u_1$. As shown in previous work (\cite{prelim_flow}), the marginal vector field generating $p_{\tau}(u_{\tau})$ is given by the following equation
\begin{equation}\begin{split}\label{eq:prelim_decomp}
    v_\tau(u_\tau) = \int v_\tau(u_\tau|u_1)\frac{p_\tau(u_\tau|u_1)p_1(u_1)}{p_\tau(u_\tau)}\d u_1.
\end{split}\end{equation}

In practice, we choose $p_0$ that is easy to sample from, e.g., Gaussian distributions, parameterize the vector field as $v^{\theta}_{\tau}$, and learn the aforementioned vector field $v_{\tau}$ by minimizing loss:
\begin{equation*}\begin{split}
    \calL_{\text{marginal}}(\theta) = \bbE_{\tau\sim\unif[0,1], u_\tau\sim p_\tau} \norm{v_\tau(u_\tau) - v^{\theta}_\tau(u_\tau)}^2.
\end{split}\end{equation*}
With \Cref{eq:prelim_decomp}, loss $\calL_{\text{marginal}}(\theta)$ simplifies to the following flow-matching loss $\calL(\theta)$ (\cite{holderrieth2025introduction}):
\begin{equation*}\begin{split}
    \calL_{\text{marginal}}(\theta) = {}& \bbE_{u_1\sim p_1, \tau\sim\unif[0,1], u_\tau\sim p_\tau(\cdot|u_1)}\\
    {}& \hspace{0em}\norm{v_\tau(u_\tau|u_1) - v^{\theta}_\tau(u_\tau)}^2 + C\\
    \eqqcolon {}& \calL(\theta) + C
\end{split}\end{equation*}
where $C$ is a constant.
This formulation translates the loss on unknown marginal vector field $v_{\tau}(u_\tau)$ to conditional vector field $v_{\tau}(u_{\tau}|u_1)$, which can be easily derived given $p_{\tau}(u_{\tau}|u_1)$, thereby making the training objective tractable. 

A popular choice of the conditional distribution is $p_\tau(\cdot|u_1) = \calN(\tau u_1,(1-\tau)^2I)$, often referred to as the conditional optimal transport (CondOT) probability path. 
It can be generated by conditional vector field $v_\tau(u_\tau|u_1) = (u_1-u_\tau)/(1-\tau)$. 
With this conditional distribution, the above flow matching loss simplifies to
\begin{equation*}\begin{split}
    {}& \calL(\theta) = \bbE_{u_1\sim p_1, \tau\sim\unif[0,1], \epsilon\sim \calN(0,I)}\\
    {}& \hspace{4em}\norm{(u_1-\epsilon) - v^{\theta}_\tau(\tau u_1+(1-\tau )\epsilon)}^2.
\end{split}\end{equation*}

\section{Training Flow-based Policies Online: From Soft Policy Iteration to Soft Actor-Critic}\label{sec:alg}
We begin this section by revisiting the Soft Policy Iteration (SPI) algorithm \cite{SAC}, which is the foundation of SAC. We then highlight the challenges that arise when incorporating flow-based policies into the SPI algorithm and introduce our approaches to tackle these challenges. 
Finally, we extend the algorithm to the SAC framework, yielding an accurate and efficient algorithm for max-entropy RL (\Cref{alg:sac}).

\subsection{SPI with Flow-based Policies}
Starting from any policy $\pi$, SPI iteratively performs soft policy evaluation and soft policy improvement. In soft policy evaluation, we estimate the soft $Q^{\pi}$ function for current policy $\pi$ by repeatedly applying the soft Bellman operator $\calT^{\pi}$ until convergence (line 5).
In soft policy improvement, we update $\pi$ to new policy $\pi^+(\cdot|x) \propto \exp\b{Q^{\pi}(x,\cdot)/\alpha}$ for every $x$ (line 6).
If every step is performed perfectly without error, the above procedure is guaranteed to learn a $\pi^\star$ such that $Q^{\pi^\star}(x,u) \geq Q^{\pi}(x,u)$ for any policy $\pi$ and $(x,u)\in\calX\times\calU$.

\begin{algorithm}[ht]
    \caption{Soft Policy Iteration \cite{SAC}}
    \label{alg:spi}
    \begin{algorithmic}[1]     
        \State \textbf{Init:} any policy $\pi^1$; 
        
        \For{episode $k=1,2,\cdots,K$}
            \State \textcolor{cyan}{// Soft Policy Evaluation}
            \State Initialize $Q^{\pi^k}(x,u)\gets 0, \forall (x,u)$;
            \State Repeat until convergence:
            \begin{equation}\begin{split}\label{eq:sac_softq}
                {}&  Q^{\pi^k}(x,u) \gets r(x,u)\\
                {}& + \gamma\bbE\left[Q^{\pi^k}(x',u') + \alpha\calH(\pi^{k}(\cdot|x'))\right], \quad \forall (x,u);
            \end{split}\end{equation}

            \State \textcolor{cyan}{// Soft Policy Improvement}
            \begin{equation}\begin{split}\label{eq:sac_update}
                \pi^{k+1}(u|x) \propto \exp\b{Q^{\pi^k}(x,u)/\alpha}, \quad \forall x;
            \end{split}\end{equation}
        \EndFor
    \end{algorithmic}
\end{algorithm}

For the rest of this section, the policy $\pi(\cdot|x)$ will be parameterized by a flow-based model with vector field $v_{x,\tau}^{\theta}(u_{x,\tau})$. We use $p_{x,\tau}^{\theta}(u_{x,\tau})$ to denote the probability generated by $v_{x,\tau}^{\theta}(u_{x,\tau})$ with $p_{x,1}^{\theta}(\cdot) = \pi(\cdot|x)$. 

To apply such flow-based policies to the SPI algorithm, the following challenges need to be resolved. 
\begin{itemize}
\item \textit{Challenge 1. How to calculate policy entropy?} The first challenge lies in policy evaluation \Cref{eq:sac_softq}, where we need to estimate the policy entropy $H(\pi(\cdot|x)) = \bbE_{u}(\log(\pi(u|x)))$. The estimation of policy entropy requires action probability $\pi(u|x)$, which is non-trivial to calculate for a flow-based policy. To tackle this issue, we will revisit the well-established technique of instantaneous change-of-variables \cite{prelim_ode_intro} (\Cref{sec:challenge1}). 
\vspace{5pt}
\item \textit{Challenge 2. How to perform policy improvement?} The second challenge lies in policy improvement \Cref{eq:sac_update}. To update the flow-based policy from $\pi(\cdot|x)$ to $\pi^+(\cdot|x)\propto \exp\b{Q^{\pi}(x,\cdot)/\alpha}$ using standard flow matching, we update its velocity field $v^{\theta}_{x,\tau}(u_{x,\tau})$ by minimizing the following loss:
\begin{equation}\begin{split}\label{eq:loss_naive}
    \calL_x(\theta) = {}& \bbE_{u\sim\pi^+(\cdot|x), \epsilon\sim \calN(0,I), \tau\sim\text{Uniform}[0,1]}\\
    {}& \hspace{0em}\norm{v^{\theta}_{x,\tau}(\tau u+(1-\tau)\epsilon) - (u-\epsilon)}^2.
\end{split}\end{equation}
This would require samples $u$ from the target distribution $\pi^+(\cdot|x)$, which are unavailable before policy improvement. To tackle this issue, we draw ideas from importance sampling \cite{importance_sampling, diffusion_1} and propose an online variant of flow matching, importance sampling flow matching (ISFM), which requires only samples from a sampling distribution with a support larger than the target distribution and is theoretically guaranteed to output a close estimation to the target distribution with enough samples (\Cref{sec:challenge2}). 
\end{itemize}

\subsection{Calculating Action Probability in Policy Evaluation}\label{sec:challenge1}
To evaluate the policy entropy in soft $Q$ function \Cref{eq:sac_softq} for any given policy $\pi(\cdot|x)$, we need to calculate action probability $\pi(u|x)$. We revisit the instantaneous change-of-variables technique for this calculation.
\begin{theorem}[Theorem 4.7 in \cite{prelim_ode_intro}]
    Consider vector field $v_{\tau}(u_\tau)$ and the generated probability $p_\tau(u_\tau)$. Suppose both of them are continuously differentiable in $(\tau,x)$. The following holds for all $\tau\in[0,1]$:
    \begin{equation*}\begin{split}
    \pushQED{\qed}
        \frac{\d \log p_{\tau}(u_\tau)}{\d \tau} = - \tr \b{\frac{\partial v_{\tau}(u_\tau)}{\partial u_\tau}}.\qedhere
        \popQED
    \end{split}\end{equation*}
\end{theorem}
Now consider any state $x$ and flow-based policy $\pi(\cdot|x)$ with velocity field $v_{x,\tau}^{\theta}(u_{x,\tau})$ and generated probability $p_{x,\tau}^{\theta}(u_{x,\tau})$. By the above theorem, we have,
\begin{equation}\begin{split}\label{eq:inst_cov}
    \log\pi(u_{x,1}|x)=& \log p_{x,1}^{\theta}(u_{x,1})\\
    =& \log p_{x,0}^{\theta}(u_{x,0}) + \int_0^1 \frac{\d \log p_{x,\tau}^{\theta}(u_{x,\tau})}{\d \tau} \d \tau\\
    =&
    \log p_{x,0}^{\theta}(u_{x,0}) - \int_0^1 \tr\b{\frac{\partial v^{\theta}_{x,\tau}(u_{x,\tau})}{\partial u_{x,\tau}}}\d \tau .
\end{split}\end{equation}
Thus, to calculate action probability $\log \pi(u_{x,1}|x))$ for action $u_{x,1} = u_{x,0} + \int_0^1 v_{x,\tau}^{\theta}(u_{x,\tau})\d \tau$, we only need to calculate Jacobian of the vector field $\partial v^{\theta}_{x,\tau}/\partial u_\tau$ and integrate the trace of the Jacobian along the path.
An empirical mean of this probability then gives an estimation of the desired entropy term $\calH(\pi(\cdot|x))$ in the soft policy evaluation \Cref{eq:sac_softq}.

\subsection{Importance Sampling Flow Matching in Policy Improvement}\label{sec:challenge2}
To perform soft policy improvement $\pi^+(\cdot|x)\propto \exp\b{Q^{\pi}(x,\cdot)/\alpha}$ for state $x$, standard flow matching requires samples from the target distribution $\pi^+(\cdot|x)$ to compute and minimize the flow matching loss \Cref{eq:loss_naive}. However, as discussed in the previous sections, such samples are generally unavailable. 

To address this issue, we adopt the idea of importance sampling by drawing samples from a easy-to-sample distribution $\tilp(\cdot|x)$, which will be specified later. 
With $\{u^{(i)}\sim\tilp(\cdot|x)\}_{i=1}^N$, we compute the following importance sampling loss:
\begin{equation}\begin{split}\label{eq:flow_matching_is}
    {}& \wh{\calL}_{\is,x}(\theta) \coloneqq \frac{1}{N}\sum_{i=1}^N \frac{\pi^+(u^{(i)}|x)}{\tilp(u^{(i)}|x)} \bbE_{\epsilon\sim \calN(0,I), \tau\sim\text{Uniform}[0,1]}\\
    {}& \hspace{2em}\norm{v^{\theta}_{x,\tau}(\tau u^{(i)}+(1-\tau)\epsilon) - (u^{(i)}-\epsilon)}^2.
\end{split}\end{equation}
In expectation, $\wh{\calL}_{\is,x}$ is equivalent to the original loss $\calL_x(\theta)$ because 
\begin{equation*}\begin{split}
    {}& \calL_{\is,x}(\theta) \coloneqq \bbE_{u^{(i)}\sim \tilp(\cdot|x)}\b{\wh{\calL}_{\is,x}(\theta)}\\
    = {}& \frac{1}{N}\sum_{i=1}^N\int\tilp(u^{(i)}|x)\frac{\pi^+(u^{(i)}|x)}{\tilp(u^{(i)}|x)}\bbE_{\epsilon\sim \calN(0,I), \tau\sim\text{Uniform}[0,1]}\\
    {}& \hspace{2em}\norm{v^{\theta}_{x,\tau}(\tau u^{(i)}+(1-\tau)\epsilon) - (u^{(i)}-\epsilon)}^2 \d u^{(i)}\\
    = {}& \int\pi^+(u|x)\bbE_{\epsilon\sim \calN(0,I), \tau\sim\text{Uniform}[0,1]}\\
    {}& \hspace{2em}\norm{v^{\theta}_{x,\tau}(\tau u+(1-\tau)\epsilon) - (u-\epsilon)}^2 \d u\\
    = {}& \calL_x(\theta).
\end{split}\end{equation*}
One can then minimize the importance sampling loss within a parameter set $\Theta$, obtain minimizer 
\begin{equation*}\begin{split}
    \wh{\theta}\coloneqq \mathop{\arg\max}_{\theta\in\Theta} \wh{\calL}_{\is,x}(\theta)\in\Theta,    
\end{split}\end{equation*}
and output policy $\wh{\pi}^+(\cdot|x) = p_{x,1}^{\wh{\theta}}(\cdot)$.

To analyze the error of $p_{x,1}^{\wh{\theta}}(\cdot)$ and quantify how the sampling distribution $\tilp(\cdot|x)$ influence this error, we present the following theorem.
\begin{theorem}\label{lem:isflow}
    Let $\Theta$ be a finite parameter space. Suppose the target distribution $\pi^+(\cdot|x)$ satisfies $\pi^+(\cdot|x) = p_{x,1}^{\theta^\star}(\cdot)$ for some  $\theta^\star\in\Theta$. Suppose $v^{\theta}_{x,\tau}(u)$ is both continuously differentiable in $(\tau,u)$ and Lipschitz in $u$ with constant $L$ for all $\theta\in\Theta$. 
    
    Let $\tilp(\cdot|x)$ be any sampling distribution whose support is a super set of the support of $p_{x,1}^{\theta^\star}(\cdot)$. Suppose 
    \begin{equation}\begin{split}\label{eq:thm1_boundedl}
        {}& \bbE_{u\sim\tilp(\cdot|x), \epsilon\sim \calN(0,I), \tau\sim\text{Uniform}[0,1]}\\
        {}& \max_{\theta\in\Theta}\norm{v^{\theta}_{x,\tau}(\tau u+(1-\tau)\epsilon) - (u-\epsilon)}^8 \leq M^2.
    \end{split}\end{equation}
    for constant $M$. Then the output distribution $p_{x,1}^{\wh{\theta}}$ satisfies
    \begin{equation*}\begin{split}
        {}&  \bbE\bs{W_2(p_{x,1}^{\theta^\star},p_{x,1}^{\wh{\theta}})^2} \leq 4\sqrt{2}\exp\b{1+2L}\\
        {}& \hspace{2em}\cdot \sqrt{M\log(2|\Theta|)}\frac{\sqrt[4]{\exp\b{3D_4(p_{x,1}^{\theta^\star}\|\tilp(\cdot|x))}}}{\sqrt{N}}.
    \end{split}\end{equation*}
    Here $W_2(p,q)$ is the 2-Wasserstein distance and $D_4(p\|q) = 1/3\ln\bbE_{u\sim q}\bs{(p(u)/q(u))^4}$ is the fourth-order Renyi divergence \cite{prelim_renyi}.
    \qed
\end{theorem}
The proof is deferred to \Cref{sec:proof}.
As shown in \Cref{lem:isflow}, when we sample from $\tilp(\cdot|x)$ and optimize the empirical loss $\wh{\calL}_{\is,x}(\theta)$, 
the error in the output distribution $p^{\widehat{\theta}}_{x,1}$ is inversely proportional to the square root of the sample size $N$ and proportional to $D_4(p_{x,1}^{\theta^\star}\|\tilp(\cdot|x))$, i.e., the fourth-order Renyi divergence between sampling distribution $\tilp(\cdot|x)$ and target distribution $p_{x,1}^{\theta^\star}$. Therefore, the closer $\tilp(\cdot|x)$ is to $p_{x,1}^{\theta^\star}$, the less samples we need for policy improvement. 

Based on this intuition, we propose to use the current policy as sampling distribution, i.e., $\tilp(\cdot|x)=\pi(\cdot|x)$. As indicated by previous works \cite{SAC,SAC_theory,maxent_lqr}, the sequence of policies $\{\pi_k\}_{k=1}^K$ from the SAC algorithm converges to an optimal policy $\pi^\star$, indicating an decreasing upper bound on the distance between current policy $\pi(\cdot|x)$ and next policy $\pi^+(\cdot|x)$. Consequently, by using $\pi(\cdot|x)$ as the sampling distribution $\tilp(\cdot|x)$, we can expect increasingly accurate policy improvement as the algorithm proceeds.

\begin{remark}
    We leave for future work to provide rigorous upper bounds on the distance, especially the fourth-order Renyi-divergence $D_4(\pi^{k+1}(\cdot|x)\|\pi^{k}(\cdot|x))$, between $\pi^k(\cdot|x)$ and $\pi^{k+1}(\cdot|x)$. Given this upper bound, one can carefully design an efficient sampling schedule that dynamically determines how many samples are needed in each episode to perform policy improvement.\qed
\end{remark}

Finally, with sampling distribution $\tilp(\cdot|x)=\pi(\cdot|x)$, we obtain the following importance sampling loss
\begin{equation}\begin{split}\label{eq:flow_matching_is_sampling}
    {}& \wh{\calL}_{\is,x}(\theta)\\
    = {}& \frac{1}{N}\sum_{i=1}^N\frac{\pi^+(u^{(i)}|x)}{\pi(u^{(i)}|x)}\bbE_{\epsilon\sim \calN(0,I), \tau\sim\text{Uniform}[0,1]}\\
    {}& \hspace{0em}\norm{v^{\theta}_{x,\tau}(\tau u^{(i)}+(1-\tau)\epsilon) - (u^{(i)}-\epsilon)}^2\\
    \propto {}& \frac{1}{N}\sum_{i=1}^N \frac{\exp\b{Q^{\pi}(x,u^{(i)})/\alpha}}{\pi(u^{(i)}|x)}\bbE_{\epsilon\sim \calN(0,I), \tau\sim\text{Uniform}[0,1]}\\
    {}& \hspace{0em}\norm{v^{\theta}_{x,\tau}(\tau u^{(i)}+(1-\tau)\epsilon) - (u^{(i)}-\epsilon)}^2.
\end{split}\end{equation}
Here in the second line we have used definition $\pi^+(\cdot|x)\propto \exp\b{Q^{\pi}(x,\cdot)/\alpha}$. Finally, to calculate $\pi(u^{(i)}|x)$, we use the instantaneous change-of-variable technique \Cref{eq:inst_cov} in the previous subsection (\Cref{sec:challenge1}).

\subsection{SAC with Flow-based Policies}
In the previous subsections, we proposed SPI (\Cref{alg:spi}) with flow-based policies. In practice, SPI performs policy evaluation and improvement for every state $x\in\calX$, which can be intractable for problems with large state space $\calX$. We now extend SPI with flow-based policies to the SAC framework \cite{SAC} to handle large and continuous state spaces using function approximation.

The algorithm maintains flow-based policy $\pi$ with vector field $v^{\theta}_{x,t}(u)$, a neural network taking $(x,t,u)$ as inputs, and $Q$ function $Q^{\psi}(x,u)$, a neural network taking $(x,u)$ as inputs. 
In policy evaluation of each episode (line 6-8), the action probability $\pi(u'|x')$ is calculated using the instantaneous change-of-variables technique \Cref{eq:inst_cov}, where the integral is estimated using second order Runge-Kutta. In policy improvement (line 9-11), we improve the policy to $\pi^+(\cdot|x)\propto \exp\b{Q^{\psi}(x,\cdot)}$ for all $x'\in\calX$ by minimizing a summation of the importance sampling losses $\sum_{x'}\widehat{\calL}_{\is,x'}(\theta)$ introduced in \Cref{eq:flow_matching_is_sampling}. 

\begin{algorithm}[h]
    \caption{SAC with Importance Sampling Flow Matching (SAC-ISFM)}
    \label{alg:sac}
    \begin{algorithmic}[1]     
        \State \textbf{Init:} policy parameter $\alpha$, $\theta$ and $\overline{\theta}$, $Q$ parameters $\psi$ and $\overline{\psi}$, and auxiliary parameters $\tau,N,K$;
        
        \For{episode $k=1,2, \cdots, K$}
            \State Sample $\{(x,u,r,x')\}$ using current policy $\pi$ and store to buffer $\calD$;
                \State Sample $\calB\gets\{(x,u,r,x')\}$ from $\calD$;
                \State For every $x'$, sample actions $\{u_{x'}^{(i)}\}_{i=1}^{N}$ from $\pi$ and compute action probabilities $\{\pi_{x'}^{(i)} = \pi(u_{x'}^{(i)}|x')\}_{i=1}^N$;
                \State \textcolor{cyan}{// Policy Evaluation}
                \State Compute policy evaluation loss:
                \begin{equation*}\begin{split}
                    {}& \calL_Q(\psi) \gets \sum_{(x,u,r,x')\in\calB}\left\|Q^{\psi}(x,u)- r\right.\\
                    {}& \left. - \gamma\sum_{i=1}^N\bs{Q^{\overline{\psi}}(x',u_{x'}^{(i)}) - \alpha\log{\pi_{x'}^{(i)}}} \right\|^2;
                \end{split}\end{equation*}
                \State Update: $\psi \gets \text{Adam}(\psi, \nabla \calL_Q(\psi)),\ \overline{\psi} \gets \tau \psi + (1-\tau)\overline{\psi}$;
                \State \textcolor{cyan}{// Policy Improvement}
                \State Compute policy improvement loss:
                \begin{equation*}\begin{split}
                    \widehat{\calL}_{\text{IS},x'}(\theta) \gets{}& \frac{1}{N}\sum_{i=1}^N\frac{\exp\b{Q^{\psi}(x',u_{x'}^{(i)})/\alpha}}{\pi_{x'}^{(i)}}\\
                    {}& \hspace{-2em}\bbE_{\epsilon,\tau}\norm{v_{x',\tau}^{\theta}(\tau u_{x'}^{(i)}+(1-\tau)\epsilon) - (u_{x'}^{(i)}-\epsilon)}^2;\\
                    \calL_{\pi}\b{\theta} \gets {}& \sum_{(x,u,r,x')\in\calB} \widehat{\calL}_{\text{IS},x'}(\theta);
                \end{split}\end{equation*}
                \State Update: $\theta \gets \text{Adam}(\theta, \nabla \calL_{\pi}(\theta))$, $\overline{\theta}\gets\tau\theta + (1-\tau)\overline{\theta}$;

        \EndFor
    \end{algorithmic}
\end{algorithm}

\section{A Case Study --- Max-Entropy LQR}\label{sec:lqr}
In this section, as a case study, we introduce the max-entropy LQR problem, a class of max-entropy RL problems with linear state transition and quadratic reward function. 
In max-entropy LQR problems, the SPI algorithm generates a policy with a closed-form solution in every episode. By analyzing the closed-loop solutions, we can theoretically confirm that the final output policy is optimal. We then perform simulations with SAC-ISFM (\Cref{alg:sac}) and compare the output policy with the optimal policy, confirming the optimality of the algorithm empirically.
\subsection{The Max-Entropy LQR Problem}
Consider the following system
\begin{equation}\begin{split}\label{eq:system}
    x^{t+1} = Ax^t + Bu^t + w^t, \quad x_0\sim\rho_0
\end{split}\end{equation}
Here $x^t\in\bbR^{d_x}, u^t\in\bbR^{d_u}$ and $w^t\in\bbR^{d_x}$ are the state, input and noise, respectively. We consider i.i.d. noises $\{w^t\}_t$ and let $\Sigma_w$ denote its second moment. $(A,B)$ are system matrices and $\rho_0$ is the initial state distribution.

For any given policy $\pi:\bbR^{d_x}\mapsto\bbR^{d_u}$, the entropy-regularized return function is defined as
\begin{equation*}\begin{split}
    J^{\pi}(x) = {}& \bbE\left[\sum_{t=0}^{\infty}\gamma^t \big[-(x^t)\t Qx^t - (u^t)\t Ru^t\right.\\
    {}& \hspace{1em} \left. + \alpha\calH(\pi(\cdot|x^t))\big] | x^0=x\right].
\end{split}\end{equation*}
Here $\gamma\in(0,1)$ is the discount factor, $Q,R\succ 0$ are unknown positive definite cost matrices. The objective is to solve for an optimal policy $\pi^\star$ that maximizes the above return. 
Throughout the paper, we consider systems satisfying the following standard assumption for $\gamma\in(0,1)$:
\begin{assumption}\label{assmp:sta}
    $(\sqrt{\gamma}A, \sqrt{\gamma}B)$ is stabilizable.
\end{assumption}

Max-entropy LQR problems is a specific class of max-entropy RL problems, where state space $\calX = \bbR^{d_x}$, action space $\calU=\bbR^{d_u}$, transition probability $\bbP$ is defined by \Cref{eq:system} and reward $r(x,u) = -(x\t Qx + u\t Ru)$. Recent literature \cite{maxent_lqr} has established an exact solution of the optimal policy in the following theorem. Mean value of the optimal poicy $-K^\star x$ is exactly the optimal controller of the standard LQR problem without entropy regularization. 
\begin{theorem}[\cite{maxent_lqr}]\label{thm:lqr_ent}
    Suppose Assumption \ref{assmp:sta} holds. The optimal policy maximizing $J^{\pi}$ is $\pi^\star(\cdot|x) = \calN(-K^\star x, \Sigma^\star), \forall x$. Here
    \begin{equation*}\begin{split}
        K^\star = {}& \gamma\b{R+B\t P^\star B}^{-1}B\t P^\star A,\\
        \Sigma^\star = {}& \frac{\alpha}{2} \b{R+\gamma B\t P^\star B}^{-1},
    \end{split}\end{equation*}
    where $P^\star$ is the solution to the riccati equation on $P$:
    \begin{equation*}\begin{split}
    \pushQED{\qed}
        P = Q + (K^\star)\t P K^\star + \gamma \b{A-BK^\star}\t P(A-BK^\star).\qedhere
        \popQED
    \end{split}\end{equation*}
\end{theorem}

\subsection{Applying SPI to Max-Entropy LQR Problems}
We now apply SPI to the max-entropy LQR problem and derive the exact solution of policy evaluation and policy improvement in every episode. 

We start the algorithm with an arbitrary Gaussian policy $\pi(\cdot|x) = \calN(-K^0x,\Sigma_u^0)$. The following lemma derives the soft Q function and the updated policy:
\begin{lemma}\label{lem:lqr}
    For any Gaussian policy $\pi(\cdot|x) = \calN(-Kx,\Sigma_u)$, the soft Q function is
    \begin{equation*}\begin{split}
        Q^{\pi}(x,u) = {}& - x\t (Q+\gamma A\t PA)x - u\t \b{R+\gamma B\t PB}u \\
        {}& - 2\gamma x\t A\t PBu - \gamma \left(c + \tr\b{P\Sigma_w}\right),
    \end{split}\end{equation*}
    where $P$ is the solution to:
    \begin{equation}\begin{split}\label{eq:lqr_riccati_withK}
        P = {}& Q + K\t RK + \gamma\b{A-BK}\t P\b{A-BK},
    \end{split}\end{equation}
    and $c$ is defined as:
    \begin{equation*}\begin{split}
        c = {}& \frac{1}{1-\gamma}\left(\tr\b{(R+\gamma B\t PB)\Sigma_u} - \alpha\left(\frac{d_u}{2}\log(2\pi e)\right.\right.\\
        {}& \left.\left.+\frac{1}{2}\log|\Sigma_u|\right) + \gamma\tr\b{P\Sigma_w}\right)\\
    \end{split}\end{equation*}
    Policy improvement on $\pi$ following \Cref{eq:sac_update} gives:
    \begin{equation*}\begin{split}
    \pushQED{\qed}
        \pi^+(\cdot|x) = \calN {}& \left(-\gamma\b{R+\gamma B\t PB}^{-1}B\t PA x,\right.\\
        {}& \left.\frac{\alpha}{2}\b{R+\gamma B\t P B}^{-1}\right). \qedhere
    \popQED
    \end{split}\end{equation*}
\end{lemma}

The above lemma shows that the policy improvement \Cref{eq:sac_update} transforms a Gaussian policy $\calN(-K^kx,\Sigma^k)$ to another Gaussian policy $\calN(-K^{k+1}x, \Sigma^{k+1})$. In the mean value, the linear controllers $K^k$ and $K^{k+1}$ follow exactly the policy iteration procedure of unregularized LQR problems.
Such a controller sequence $\{K^k\}$ from policy iteration is known to converge to $K^\star$ in \Cref{thm:lqr_ent} (\cite{lqr_policy_iteration}). Furthermore, with $K=K^\star$ and $P=P^\star$, the corresponding covariance $\Sigma=\alpha\b{R+\gamma B\t PB}^{-1}/2$ equals $\Sigma^\star$ in \Cref{thm:lqr_ent}. 
We can then conclude that applying SPI to max-entropy LQR problems outputs the optimal policy derived in \Cref{thm:lqr_ent}.

\subsection{Applying SAC-ISFM to Max-Entropy LQR Problems}
Consider a max-entropy LQR problem with the following system parameters:
\begin{equation}\begin{split}\label{eq:lqrsys}
    A = {}& \begin{bmatrix}
        0.55 & 0.55 & 0 & 0 & 0\\
        0.0 & 0.55 & 0.55 & 0 & 0\\
        0.0 & 0.0 & 0.55 & 0.55 & 0\\
        0.0 & 0.0 & 0.0 & 0.55 & 0.55\\
        0.55 & 0.0 & 0.0 & 0.0 & 0.55
    \end{bmatrix}, \quad B = I_5,\\
    Q = {}& R = I_5, \quad \gamma = 0.9, \quad \Sigma_w = I_5, \quad x_0 = 0.
\end{split}\end{equation}

\begin{figure}[H]
    \centering
    \includegraphics[width=0.45\textwidth]{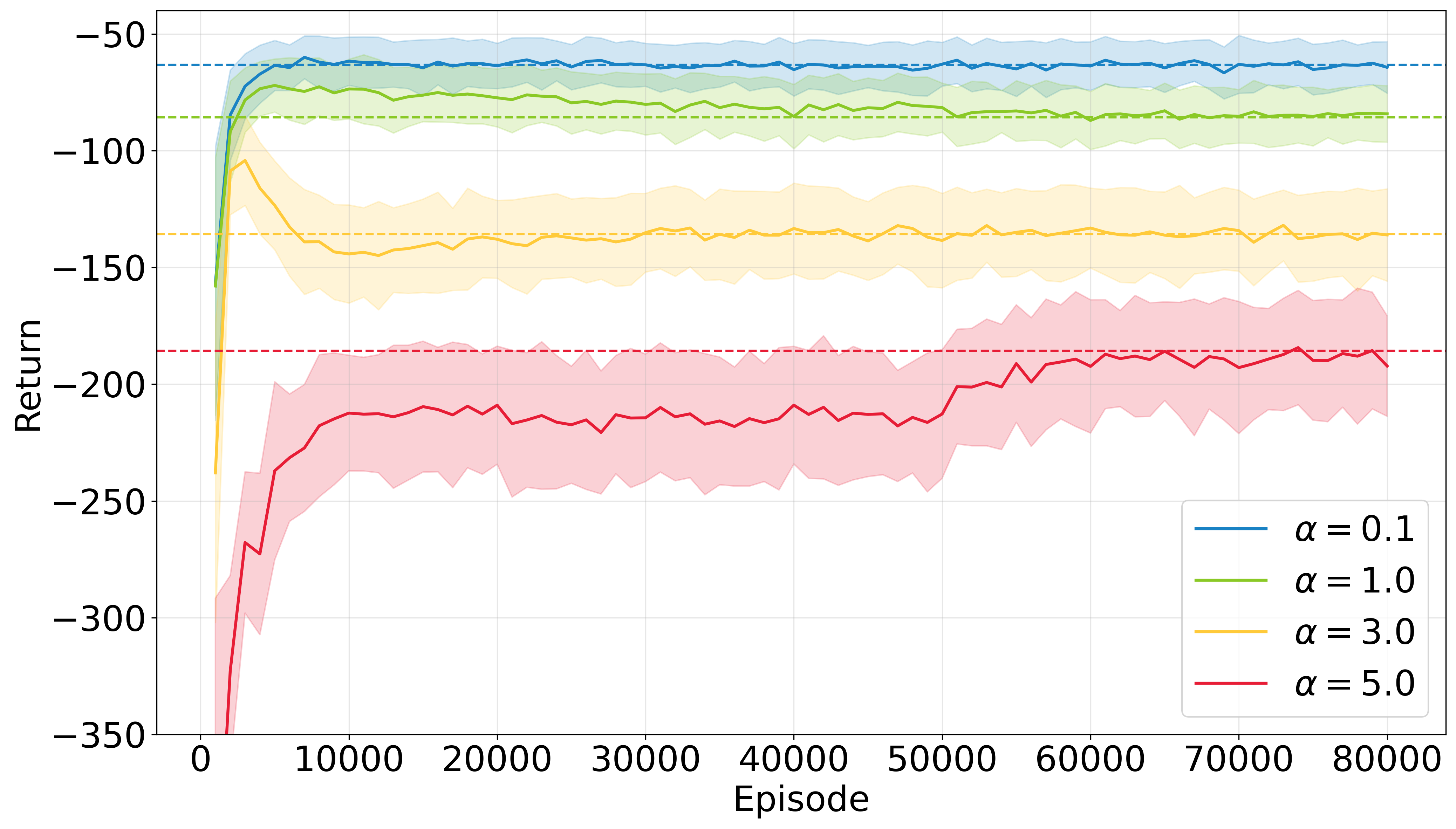}
    \caption{Discounted LQR return (without entropy regularization) during training for different $\alpha$.}    
    \label{fig:training}
\end{figure}

We ran \Cref{alg:sac} for each $\alpha\in\{0.1, 1, 3, 5\}$. For the trained policy $\pi$, we estimated the standard LQR returns without entropy regularization, averaged over 100 trajectories, and presented them as solid lines. We also present the returns of the optimal controller, derived in \Cref{lem:lqr}, as dashed lines. 
As shown in \Cref{fig:training}, the return converged to the optimal value for each $\alpha$.

\begin{figure}[H]
    \centering
    \includegraphics[width=0.45\textwidth]{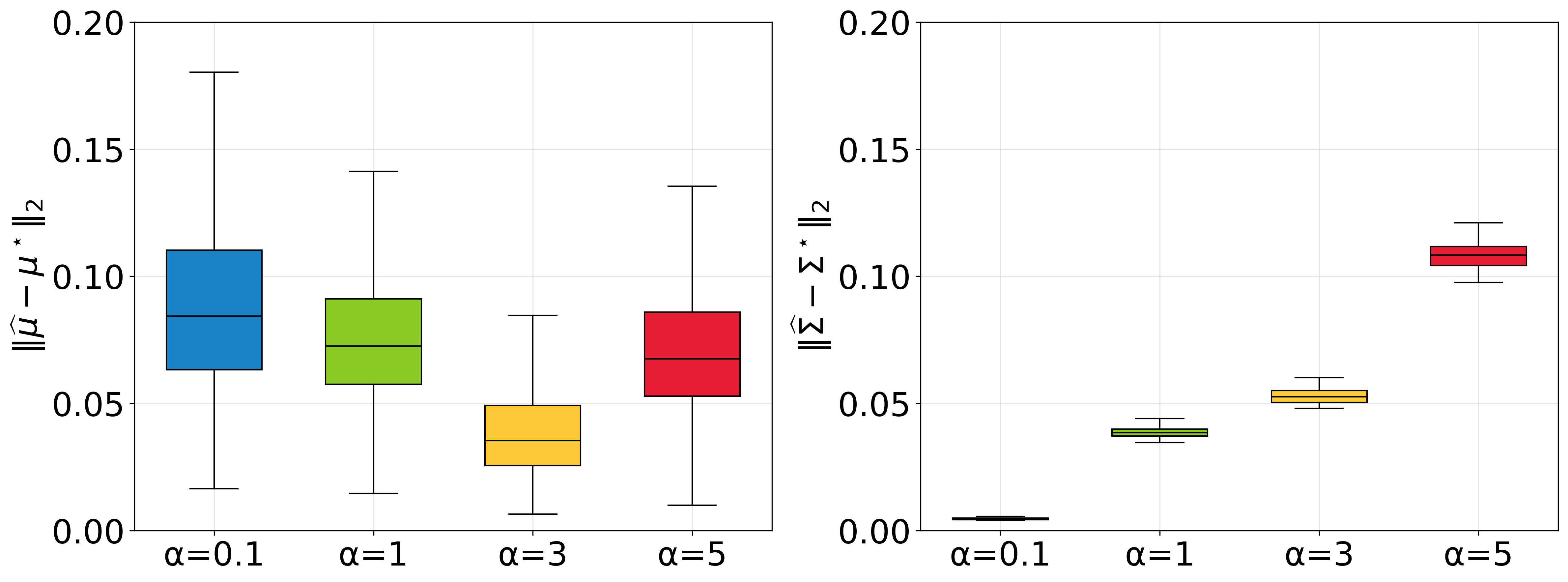}
    \label{fig:final_policy}
    \caption{Mean differences and covariances difference between $\pi$ in episode $K=80000$ and the optimal policy (\Cref{thm:lqr_ent}), averaged across states from $50$ trajectories with length 100.}
\end{figure}

We further evaluated the policy $\pi$ in episode $K=80000$ for each $\alpha\in\{0.1, 1, 3, 5\}$. Specifically, for each $\alpha$, we sampled $50$ state trajectories with length $100$ using the output policy. For each state in the trajectories, we sampled $12800$ actions, estimated the mean $\widehat{\mu}$ and covariance $\widehat{\Sigma}$, and calculated the distance to the optimal mean $\mu^\star$ and covariance $\Sigma^\star$ (\Cref{thm:lqr_ent}) in $2$-norm. We plotted the average and std of $\|\widehat{\mu}-\mu^\star\|_2$ and $\|\widehat{\Sigma}-\Sigma^\star\|_2$ across all states in  the trajectories for each alpha. As shown in \Cref{fig:final_policy}, both $\widehat{\mu}$ and $\widehat{\Sigma}$ are close to the optimal value in 2-norm. 


\section{Conclusion}
In this paper, we propose variants of the SPI and SAC algorithms with flow-based policies. In policy evaluation, we employ the instantaneous change-of-variables to estimate entropy of the flow-based policy. In policy improvement, we propose importance sampling flow matching, a provably efficient algorithm that updates the flow-based policy using samples from a user-specified sampling distribution rather than the target distribution as required by standard flow matching. 
Finally, we conduct a case study of the proposed algorithms on max-entropy LQR problems. We confirm that SPI and SAC with flow-based policies learn the optimal solution accurately.




\bibliography{ref}

\appendices

\section{Proofs}\label{sec:proof}
\subsection{Proof of \Cref{lem:isflow}}
\begin{proof}
    We begin by introducing necessary notations. For simplicity, we omit all dependency $x$ in this proof. 
    Recall that $p_1^{\theta^\star}(u)$ is the target distribution, $\tilp(u)$ is the sampling distribution, and $v_{\tau}^{\theta^\star}(u)$ is the velocity field generating $p_{\tau}^{\theta^\star}(u)$. We also define
    \begin{equation}\begin{split}
        w(u) \coloneqq {}&  p_1^{\theta^\star}(u)/\tilp(u)\\
        l^{\theta}(u) \coloneqq {}& \bbE_{\epsilon\sim \calN(0,I), \tau\sim\text{Uniform}[0,1]}\\
        {}& \norm{v^{\theta}_{\tau}(\tau u+(1-\tau)\epsilon) - (u-\epsilon)}^2.
    \end{split}\end{equation}
    Moreover, let
    \begin{equation}\begin{split}\label{eq:L}
        {}& \calL(\theta) \coloneqq \bbE_{u\sim\tilp}[w(u)l^{\theta}(u)] = \bbE_{u\sim p_1^{\theta^\star}} [l^{\theta}(u)],\\
        {}& \wh{\calL}(\theta;\{u^{(i)}\}_{i=1}^N) \coloneqq \frac{1}{N}\sum_{i=1}^N w(u^{(i)})l^{\theta}(u^{(i)}),\\
        {}& \wh{\theta}\b{\{u^{(i)}\}_{i=1}^N} \coloneqq \mathop{\arg\min}_{\theta\in\Theta} \wh{\calL}(\theta;\{u^{(i)}\}_{i=1}^N).
    \end{split}\end{equation}
    Whenever it's clear from the context, we omit the dependences of $\wh{\calL}$ and $\wh{\theta}$ on $\{u^{(i)}\}_{i=1}^N$.

    To prove the theorem, we first establish an upper bound on $\bbE_{u^{(i)}\sim\tilp}\calL\b{\wh{\theta}\b{\{u^{(i)}\}_{i=1}^N}}$. Then we connect this expected loss to the expected deviation of learned distribution $p_1^{\wh{\theta}}$ from the target distribution $p_1^{\theta^\star}$.

    \textbf{Step 1: Upper bound on $\bm{\bbE_{u^{(i)}\sim\tilp}\calL\b{\wh{\theta}\b{\{u^{(i)}\}_{i=1}^N}}}$.} First, consider a fixed set of data points $\{u^{(i)}\}_{i=1}^N$. By definition of $\wh{\theta}$, we know that $\wh{\calL}(\wh{\theta})-\wh{\calL}(\theta^\star) \leq 0$. Therefore,
    \begin{equation}\begin{split}
        \calL(\wh{\theta})-\calL(\theta^\star) \leq {}& \calL(\wh{\theta}) - \wh{\calL}(\wh{\theta}) + \wh{\calL}(\wh{\theta})-\wh{\calL}(\theta^\star)\\
        {}& \hspace{2em} + \wh{\calL}(\theta^\star) - \calL(\theta^\star)\\
        \leq {}& \calL(\wh{\theta}) - \wh{\calL}(\wh{\theta}) + \wh{\calL}(\theta^\star) - \calL(\theta^\star)\\
        \leq {}& 2\max_{\theta\in\Theta}\norm{\calL(\theta)-\wh{\calL}(\theta)}.
    \end{split}\end{equation}
    This leads to 
    \begin{equation}\begin{split}\label{eq:EL}
        {}& \bbE\bs{\calL(\wh{\theta})} \leq \calL(\theta^\star)\\
        + {}& \frac{2}{N} \underbrace{\bbE_{u^{(i)}\sim\tilp}\bs{\max_{\theta\in\Theta} \norm{\sum_{i=1}^N w(u^{(i)})l^{\theta}(u^{(i)}) - \bbE_{u\sim\tilp} [w(u)l^{\theta}(u)]}}}_{\calI_1}.
    \end{split}\end{equation}

    Consider the last term $\calI_1$. By the standard symmetrization lemma (Lemma 11.4 in \cite{symmetrization} \footnote{We use the lemma by letting $s=\theta$ and $x_{i,s}=w(u^{(i)})l^{\theta}(u)$. We use the first equation in the proof, omitting the very first equality.}), we know that 
    \begin{equation}\begin{split}
        \calI_1 \leq {}& 2\bbE_{u^{(i)}\sim\tilp, \sigma_i\in\unif\{\pm 1\}}\bs{\max_{\theta\in\Theta} \norm{\sum_{i=1}^N \sigma_iw(u^{(i)})l^{\theta}(u^{(i)})}}.
    \end{split}\end{equation}
    Here we have introduced auxiliary random variables $\sigma_i\sim\unif\{\pm1\}$. The RHS is exactly the Rademacher complexity of function class $\calF=\{f^{\theta}(\cdot)=w(\cdot)l^{\theta}(\cdot), \forall \theta\in\Theta\}$. Now we define $\calF' = \{\pm f, \forall f\in\calF\}$, and upper bound $\calI_1$ as follows
    \begin{equation}\begin{split}\label{eq:i1}
        \calI_1 \leq {}& 2\bbE_{u^{(i)}\sim\tilp, \sigma_i\in\unif\{\pm 1\}}\bs{\max_{f\in\calF} \norm{\sum_{i=1}^N \sigma_if(u^{(i)})}}\\
        \leq {}& 2\bbE_{u^{(i)}\sim\tilp, \sigma_i\in\unif\{\pm 1\}}\bs{\max_{f\in\calF'} \norm{\sum_{i=1}^N \sigma_if(u^{(i)})}}\\
        = {}& 2\bbE_{u^{(i)}\sim\tilp}\underbrace{\bbE_{\sigma_i\in\unif\{\pm 1\}}\bs{\max_{f\in\calF'} \sum_{i=1}^N \sigma_if(u^{(i)})}}_{\calI_2}\\
    \end{split}\end{equation}
    We then upper bound $\calI_2$ by the well established Massart lemma (Theorem 3.7 in \cite{mohri2018foundations} \footnote{We use the theorem by letting $\calA=\calF'$ and $x_i=f(u^{(i)})$.}), 
    \begin{equation}\begin{split}
        \calI_2 \leq {}& \max_{f\in\calF'} \sqrt{\sum_{i=1}^N (f(u^{(i)}))^2} \sqrt{2\log(|\calF'|)}\\
        = {}& \max_{\theta\in\Theta}\sqrt{\sum_{i=1}^N (w(u^{(i)})l^{\theta}(u^{(i)}))^2} \sqrt{2\log(2|\Theta|)}\\
        \leq {}& \sqrt{\sum_{i=1}^N (w(u^{(i)})\max_{\theta\in\Theta}l^{\theta}(u^{(i)}))^2} \sqrt{2\log(2|\Theta|)}\\
    \end{split}\end{equation}
    Substituting back into \Cref{eq:i1} gives
    \begin{equation}\begin{split}
        \calI_1 \leq {}& 2\sqrt{2\log(2|\Theta|)}\cdot \bbE_{u^{(i)}\sim\tilp}\sqrt{\sum_{i=1}^N (w(u^{(i)})\max_{\theta\in\Theta}l^{\theta}(u^{(i)}))^2}\\
        \leq {}& 2\sqrt{2\log(2|\Theta|)} \sqrt{N\bbE_{u\sim\tilp}\bs{w^2(u)\b{\max_{\theta\in\Theta}l^{\theta}(u)}^2}}\\
        \leq {}& 2\sqrt{2N\log(2|\Theta|)}\\
        {}& \cdot \sqrt[4]{\bbE_{u\sim\tilp}\bs{w^4(u)}\bbE_{u\sim\tilp}\bs{\b{\max_{\theta\in\Theta}l^{\theta}(u)}^4}}\\
        \leq {}& 2\sqrt{2MN\log(2|\Theta|)} \sqrt[4]{\bbE_{u\sim\tilp}\bs{w^4(u)}}\\
    \end{split}\end{equation}
    Here the last line is by \Cref{eq:thm1_boundedl}. Now by definition $D_{4}(p_1^{\theta^\star}\|\tilp) = 1/3\ln\bbE_{u\sim\tilp}\bs{(p_1^{\theta^\star}(u)/\tilp(u))^4}$, we have
    \begin{equation}\begin{split}
        \calI_1 \leq 2\sqrt{2MN\log(2|\Theta|)}\sqrt[4]{\exp\b{3D_4(p_1^{\theta^\star}\|\tilp)}}.
    \end{split}\end{equation}
    Finally, combining with \Cref{eq:EL} gives
    \begin{equation}\begin{split}\label{eq:ELloss}
        {}& \bbE\bs{\calL(\wh{\theta})-\calL(\theta^\star)}\\
        \leq {}& 4\sqrt{2M\log(2|\Theta|)}\frac{\sqrt[4]{\exp\b{3D_4(p_1^{\theta^\star}\|\tilp)}}}{\sqrt{N}}.
    \end{split}\end{equation}

    \textbf{Step 2: Upper bound on the deviation of $p_1^{\wh{\theta}}$ from $p_1^{\theta^\star}$. } 
    By Proposition 3 in \cite{albergo2022building}, we know that 
    \begin{equation}\begin{split}\label{eq:dist_marginal}
        {}&  \b{W_2(p_1^{\theta^\star},p_1^{\wh{\theta}})}^2 \leq \exp\b{1+2L}\\
        {}&\cdot \underbrace{\bbE_{u\sim p_1^{\theta^\star}, t\in\unif[0,1], u_t\sim p_t^{\theta^\star}(\cdot|u)}\norm{v^{\wh{\theta}}_t(u_t)-v^{\theta^\star}_t(u_t)}^2}_{\calL_{\text{marginal}}(\wh{\theta})}.
    \end{split}\end{equation}
    To connect the above loss on the marginal vector field $v^{\theta^\star}_t(u_t)$ to loss \Cref{eq:L} on the conditional vector field $v^{\theta^\star}_t(u_t|u)$, we refer to Theorem 18 in \cite{holderrieth2025introduction} stating that 
    \begin{equation}\begin{split}
        {}& \calL_{\text{marginal}}(\theta) = \calL(\theta) + C\\
    \end{split}\end{equation}
    for any $\theta$ and some constant $C$ independent of $\theta$. Therefore,
    \begin{equation}\begin{split}
        {}& \bbE\bs{\calL_{\text{marginal}}(\wh{\theta}) - \calL_{\text{marginal}}(\theta^\star)} = \bbE\bs{\calL(\wh{\theta}) - \calL(\theta^\star)}\\
        \leq {}& 4\sqrt{2M\log(2|\Theta|)}\frac{\sqrt[4]{\exp\b{3D_4(p_1^{\theta^\star}\|\tilp)}}}{\sqrt{N}}\\
    \end{split}\end{equation}
    Substituting back into \Cref{eq:dist_marginal}, we can now conclude
    \begin{equation}\begin{split}
        {}&  \bbE\bs{W_2(p_1^{\theta^\star},p_1^{\wh{\theta}})^2} \leq \exp\b{1+2L}\\
        {}& \cdot \b{\calL_{\text{marginal}}(\theta^\star) + 4\sqrt{2M\log(2|\Theta|)}\frac{\sqrt[4]{\exp\b{3D_4(p_1^{\theta^\star}\|\tilp)}}}{\sqrt{N}}}.
    \end{split}\end{equation}
    Utilizing the fact that $\calL_{\text{marginal}}(\theta^\star)=0$, we obtain the desired result.
\end{proof}

\subsection{Proof of \Cref{lem:lqr}}
\begin{proof}[Proof of \Cref{lem:lqr}]
    \textbf{Step 1: Deriving the soft Q function.}
    We begin by arguing $J^{\pi}(x) = -x\t Px - c$ for some positive definite matrix $P$ and some constant $c$. To do this, we first rewrite $J^{\pi}(x)$ as follows
    \begin{equation*}\begin{split}
        {}&  J^{\pi}(x)\\
        = {}& \sum_{t=0}^{\infty}\gamma^t \bbE\bs{-(x^t)\t Qx^t - (u^t)\t Ru^t + \alpha\calH(\pi(\cdot|x^t))}\\
        = {}& \sum_{t=0}^{\infty}\gamma^t \bbE\bs{-(x^t)\t Qx^t - (u^t)\t Ru^t} + \sum_{t=0}^{\infty}\gamma^t \bbE\bs{\alpha\calH(\pi(\cdot|x^t))}.
    \end{split}\end{equation*}
    Here the first term can be seen as the value function for a standard LQR problem with deterministic controller $u=-Kx$ and zero-mean process noises $\{B\b{u^t+Kx^t}+w^t\}_t$. Therefore it can be written as $-x\t P x + c_1$ for some positive definite matrix $P$ and constant $c_1$. The second term in $J^{\pi}(x)$ is a constant independent of $x$ because $\pi(\cdot|x)$ is a gaussian distribution with constant covariance $\Sigma_u$ and $\calH(\pi\b{\cdot|x^t}) = d_u\log(2\pi e)/2+\log|\Sigma_u|/2$. Therefore, $J^{\pi}(x) = -x\t P x - c$.

    Now we calculate $P$ and $c$. We first express $J^{\pi}(x)$ recursively as follows:
    \begin{equation*}\begin{split}
        {}& J^{\pi}(x) = -x\t P x - c\\
        = {}& \bbE\Bigg[\sum_{t=0}^{\infty} \gamma^t\b{-(x^t)\t Qx^t - (u^t)\t R u^t + \alpha\calH\b{\pi(\cdot|x^t)}}\\
        {}& \hspace{2em}\Big|x^0=x\Bigg]\\
        = {}& -x\t \b{Q+K\t RK}x - \tr\b{R\Sigma_u} + \alpha\calH(\pi\b{\cdot|x})\\
        {}& + \gamma\bbE_{x^1}\b{J^{\pi}(x^1)}\\
        = {}& -x\t \b{Q+K\t RK}x - \tr\b{R\Sigma_u} + \alpha\calH(\pi\b{\cdot|x})\\
        {}& - \gamma\bbE_u\bs{(Ax+Bu)\t P (Ax+Bu)} - \gamma\tr\b{P\Sigma_w} - \gamma c\\
        = {}& -x\t \b{Q+K\t RK + \gamma\b{A-BK}\t P\b{A-BK}}x\\
        {}& - \tr\b{\b{R+\gamma B\t PB}\Sigma_u} + \alpha\calH(\pi\b{\cdot|x})\\
        {}& - \gamma\tr\b{P\Sigma_w} - \gamma c.
    \end{split}\end{equation*}
    Here in the second last equation we used $J^{\pi}(x^1) = -(x^1)\t P x^1 - c$.
    By comparing the coefficients between the first and last line, we get
    \begin{equation*}\begin{split}
        c = {}& \frac{1}{1-\gamma}\left(\tr\b{\b{R+\gamma B\t PB}\Sigma_u} - \alpha\calH(\pi\b{\cdot|x})\right.\\
        {}& \left.+ \gamma\tr\b{P\Sigma_w}\right)\\
        = {}& \frac{1}{1-\gamma}\left(\tr\b{(R+\gamma B\t PB)\Sigma_u} - \alpha\left(\frac{d_u}{2}\log(2\pi e)\right.\right.\\
        {}& \left.\left.+\frac{1}{2}\log|\Sigma_u|\right) + \gamma\tr\b{P\Sigma_w}\right)\\
        P = {}& Q + K\t RK + \gamma\b{A-BK}\t P\b{A-BK}.
    \end{split}\end{equation*}
    Finally, by the definition of $Q^{\pi}$, we know that
    \begin{equation*}\begin{split}
        {}& Q^{\pi}(x,u)\\
        = {}&  -x\t Qx - u\t Ru + \gamma \bbE_{x'}{J^{\pi}(x')}\\
        = {}&  - x\t Qx - u\t Ru\\
        {}& + \gamma \left(-\b{Ax+Bu}\t P\b{Ax+Bu} - c - \tr\b{P\Sigma_w}\right)\\
        = {}&  - x\t (Q+\gamma A\t PA)x - u\t \b{R+\gamma B\t PB}u \\
        {}& - 2\gamma x\t A\t PBu - \gamma \left(c + \tr\b{P\Sigma_w}\right)\\
    \end{split}\end{equation*}

    \textbf{Step 2: Policy Improvement.} By \Cref{eq:sac_update},
    \begin{equation*}\begin{split}
        {}& \pi^+(u|x) \propto \exp\b{Q^{\pi}(x,u)/\alpha}\\
        \propto {}& \exp\b{\frac{1}{\alpha}\b{-u\t \b{R+\gamma B\t PB}u- 2\gamma x\t A\t PBu}}\\
        \propto {}& \exp\left(-\frac{1}{\alpha}\b{u+\gamma\b{R+\gamma B\t PB}^{-1}B\t PA x}\t\right.\\
        {}& \left.\b{R+\gamma B\t PB}\b{u+\gamma\b{R+\gamma B\t PB}^{-1}B\t PA x}\right)\\
        = {}& \calN\left(-\gamma\b{R+\gamma B\t PB}^{-1}B\t PA x,\right.\\
        {}& \left.\frac{\alpha}{2}\b{R+\gamma B\t P B}^{-1}\right).
    \end{split}\end{equation*}
\end{proof}

\end{document}